\documentclass[runningheads]{llncs}
\usepackage{graphicx}
\usepackage{amsmath,amssymb} % define this before the line numbering.
\usepackage{color}
\usepackage{booktabs}
\usepackage[width=122mm,left=12mm,paperwidth=146mm,height=193mm,top=12mm,paperheight=217mm]{geometry}

\newcommand{\ra}[1]{\renewcommand{\arraystretch}{#1}}

\begin{document}
\pagestyle{headings}
\mainmatter

\title{Going Further with Point Pair Features.} % Replace with your title
\titlerunning{Going Further with Point Pair Features}
\authorrunning{Stefan Hinterstoisser, Vincent Lepetit, Naresh Rajkumar, Kurt Konolige}
\author{Stefan Hinterstoisser$^1$, Vincent Lepetit$^2$, Naresh Rajkumar$^1$, Kurt Konolige$^1$}

%Please write out author names in full in the paper, i.e. full given and family names. 
%If any authors have names that can be parsed into FirstName LastName in multiple ways, please include the correct parsing, in a comment to the volume editors:
%\index{Lastnames, Firstnames}
%(Do not uncomment it, because you may introduce extra index items if you do that...)

\institute{\small$^1$Google,\\
	   \small\email{\{hinterst,nareshkumar,konolige\}@google.com} \\
           \small$^2$TU-Graz,\\
	   \small\email{\{lepetit\}@icg.tugraz.at }
}

\maketitle

\begin{abstract}
  Point Pair  Features is  a widely used  method to detect  3D objects  in point
  clouds,  however they  are  prone to  fail  in presence  of  sensor noise  and
  background  clutter.  We  introduce  novel sampling  and  voting schemes  that
  significantly  reduces  the  influence  of  clutter  and  sensor  noise.   Our
  experiments show that  with our improvements, PPFs  become competitive against
  state-of-the-art  methods  as it  outperforms  them  on several  objects  from
  challenging benchmarks, at a low computational cost.
\end{abstract}

\section{Introduction}
Object  instance recognition  and  3D pose  estimation have  received  a lot  of
attention   recently, probably  because  of   their  importance   in  robotics
applications~\cite{1,6,7,8,10,23,25,9}.   For grasping  and manipulation  tasks,
efficiency,  reliability, and  accuracy are  all desirable  properties that  are
still very challenging in general environments.

%% ,  when  compared  to  class
%% recognition, may  look like  an easy  task especially in  the context  of recent
%% progress made  with deep learning.   However, especially in a  robotics context,
%% methods are  usually expected to  be much  more accurate, precise  and efficient
%% than current results in object class recognition suggest.

%% For instance, if robots try to perform grasping and manipulation tasks, reliable
%% and efficient object instance recognition and pose estimation is necessary to
%% perform the task in reasonable time and to avoid harming humans, animals or
%% other objects by misinterpreting objects through false positives. Thus, the
%% ultimate goal is to design a system that is efficient, robust and highly
%% accurate and works well for all sort of objects in challenging real-world
%% settings, such as cluttered environments with heavy occlusion and lighting
%% changes.

While many approaches have been developed over  the last years, we focus here on
the approach from  Drost et al.~\cite{1}, which relies on  a depth camera.  Since
its   publication,    it   has    been   improved    and   extended    by   many
authors~\cite{6,2,3,4,5}.   However, we  believe  it has  not  delivered its  full
potential yet: Drost's technique and its variants are based on votes from pairs of 3D points, but the sampling of these pairs has been overlooked so far. As a result, these techniques are very inefficient, and
it usually still takes several seconds to run them.

Moreover, this approach is also very sensitive to sensor noise and 3D background
clutter---especially if it is close  to the target object: Sensor noise
can  disturb the  quantization on  which the  approach relies  heavily for  fast
accesses. Background clutter  casts spurious votes that can mask  the effects of
correct ones.  As a result,  several other approaches~\cite{7,8,10,9} have shown
significantly better performance on recent datasets~\cite{7,9}.

In this  paper, we propose a  much better and efficient  sampling strategy that,
together with small  modifications to the pre- and  post-processing steps, makes
our  approach competitive  against state-of-the-art  methods: It  beats them  on
several objects on recent challenging datasets, at a low computational cost.

%% We call our method EPPF, for Efficient Point Pair Feature.

In the remainder of this paper, we  review related work, and describe in
detail  the  method  of  Drost  et  al.~\cite{1}.  We  then  introduce  our
contributions, which we compare against state-of-the-art methods in the last section.

\begin{figure*}[t]
\begin{center}
\includegraphics[clip,width=0.75\linewidth]{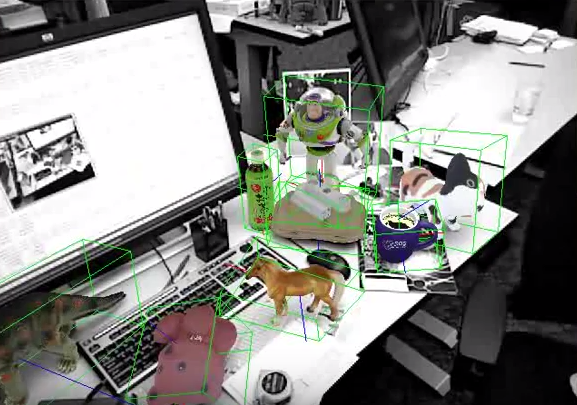}
\end{center}
\caption{\label{fig:teaser1} Several 3D objects are simultaneously detected with
  our  method  under  different  poses  on  cluttered  background  with  partial
  occlusion and  illumination changes.  Each  detected object is  augmented with
  its 3D  model, its  3D bounding  box and its  coordinate systems.   For better
  visibility, the  background is kept in  gray and only detected  objects are in
  color.}
\end{figure*}

\section{Related Work}

The literature  on object  detection and  3D pose estimation  is very  broad. We
focus here on recent work only and split them in several categories.

\paragraph{Sparse Feature-Based Methods.} While popular for 3D object
detection in color or intensity images several years  ago, these methods are less popular now
as practical robotics applications often consider  objects that do not  exhibit many
stable feature points due to the lack of texture.

\paragraph{Template-based methods.}
Several              methods              are              based              on
templates~\cite{7,8,Wohlhart15}, where the templates
capture the  different appearances  of objects  under different  viewpoints.  An
object is detected when a template matches the image and its 3D pose is given by
the template. \cite{7} uses synthetic  renderings of a 3D object
model to generate a large number of templates covering the full view hemisphere.
It  employs an  edge-based distance  metric  which works  well for  textureless
objects, and  refines the pose  estimates using ICP  to achieve an  accurate 6D
pose.   Such  template-based  approaches  can work  accurately  and  quickly  in
practice.   However, they  show typical  problems such  as not  being robust  to
clutter and occlusions.

\paragraph{Local patch-based methods.}
\cite{10,Tejani2014} use forest-based voting schemes on local patches
to detect and estimate 3D poses.   While the former regresses object coordinates
and conducts  a subsequent  energy-based pose estimation,  the latter  bases its
voting on  a scale-invariant  patch representation  and returns
location and pose simultaneously.    \cite{Bonde2014} also uses Random Forests to
infer object  and pose,  but  via  a sliding window  through a  depth volume.
In the experiment section, we compare our method against the recent approach of~\cite{10},
which performs very well.

%% These methods remain slow, and it is not clear how they scale in performance
%% with the number of objects.

\paragraph{Point-cloud-based methods.} Detection of 3D objects  in point cloud data has a
very long  history.  A review  can be found  in~\cite{14}.  One of  the standard
approaches for object  pose estimation is ICP~\cite{16},  which however requires
an initial estimate and is not suited for object detection.  Approaches based on
3D features  are more  suitable and  are usually  followed by  ICP for  the pose
refinement.  These methods include point pairs~\cite{1,16}, spin-images~\cite{15}, 
and   point-pair   histograms~\cite{17,18}.   These   methods   are   usually
computationally expensive, and have difficulty in scenes with heavy clutter. However, we 
show in this paper that these  drawbacks can be avoided.

The point cloud-based method proposed in~\cite{1} is the starting point of
our own method, and we detail it in the next section.

\section{``Drost-PPF''~\cite{1}}
\label{DrostSec}
One  of the  most promising  algorithms based  on point  clouds for  matching 3D
models to 3D scenes  was proposed by Drost et al.~\cite{1}---in  the rest of the
paper, we will  refer to it as \emph{Drost-PPF}. Since our approach is 
based on it, we will describe it here in detail.

\newcommand{\bF}{{\bf F}}
\newcommand{\ba}{{\bf a}}
\newcommand{\bb}{{\bf b}}
\newcommand{\bd}{{\bf d}}
\newcommand{\bm}{{\bf m}}
\newcommand{\bn}{{\bf n}}
\newcommand{\bs}{{\bf s}}
\newcommand{\br}{{\bf r}}

The authors of Drost-PPF coupled the  existing idea of Point Pair Features~(PPF)
with a voting scheme to solve  for the object pose and location simultaneously.
More specifically, the goal of Drost-PPF is to establish for each scene point a correspondence to a model point and solve for the remaining degree of freedom to align the scene with the model data.
This is done by using the relations between the corresponding points and all their neighboring points in model and scene space to vote in a Hough transform fashion for both unknowns.

Drost-PPF begins by extracting pairs of 3D points and their normals from the object's 3D model, to compute Point Pair Features (PPF) defined as a 4-vector:
\begin{equation}
\bF(\bm_1, \bm_2, \bn_1, \bn_2) = \left[\|\bd\|_2, \angle(\bn_1, \bd),
  \angle(\bn_2, \bd), \angle(\bn_1, \bn_2) \right]^\top \> ,
\label{eq:PPF}
\end{equation}
where  $\bm_1$ and  $\bm_2$ are  two  3D points  and $\bn_1$  and $\bn_2$  their
normals, $\bd  = \bm_2-\bm_1$, and  $\angle(\ba,\bb)$ the angle  between vectors
$\ba$ and $\bb$. This feature vector is invariant to rigid motions, which allows
the method to detect the objects under these transformations.

The PPFs are discretized and used as indices of a lookup table. Each bin of this
lookup table  stores the  list of  first model points $\bm_i$ 
and the corresponding rotation angles ${\alpha_i}^{model}$ of all the PPFs
that are discretized  to the bin's index. In this context, ${\alpha_i}^{model}$ describes the rotation angle
around the normal of the first model point $\bm_i$ that aligns the corresponding point pair that is descretized to the bin's index to a fixed canonical frame---as described in~\cite{1}.   
The stored data will  be used during the voting stage.

%% The rotation angle is  the angle of the rotation around the  normal of the first
%% model point such that the point pair  is aligned in a fixed canonical coordinate
%% frame with the first  model point lying at the center  of this coordinate frame,
%% its normal being  aligned with its x-axis  and the projection of  $\bd$ onto the
%% y-z plane being  aligned with the y-axis.

At run-time, Drost-PPF tries to establish for each scene point a correspondence to a model point.
This is achieved by pairing each scene point with all other scene points, compute the corresponding
PPFs and match them with similar PPFs computed from 3D points on the target  objects. The latter step can be performed efficiently using the lookup table: The content of the bins of the lookup table indexed by the discretzed PPFs are used to vote for pairs made of model points and discretized rotation angles $\alpha_i = {\alpha_i}^{model} - \alpha^{scene}$ in a Hough transform fashion. In this context, $\alpha^{scene}$ is the rotation around the normal of the first scene point that aligns the current scene point pair with the fixed canonical frame. 
Once a peak in the Hough space is extracted, the 3D pose of the object can be easily estimated from the
extracted scene and model point correspondence and its corresponding rotation $\alpha_i$.

%Peaks in the Hough space correspond to potential pairs of point correspondences along with its rotation.
%As shown by~\cite{1}, $\alpha_i = {\alpha_i}^{model} - \alpha^{scene}$ can be split into a model and a scene part and describes the rotation around the aligned normal of a scene and a model point pair. 
%The 3D pose of the object can then be estimated easily from the extracted scene and model point correspondence and their rotation around the aligned normals. 
%the which is extracted by a peak in the accumulator space of the Hough transform.

%The 3D pose  of the object,  assuming the match \stefan{between the scene and model point pair} is correct, can  be estimated easily from the 3D  points and their normals up  to one  rotation $\alpha$ along the  first  normal. \stefan{The rotation $\alpha$ needs to be computed for each scene and model point pair and, as shown in~\cite{1}, can be obtained by $\alpha = {\alpha}^{model} - \alpha^{scene}$, where $\alpha^{scene}$ is the rotation angle around the normal of the first scene point to align the scene point pair to a canonical frame.}
%The  rotation  angles $\alpha$ are are  also discretized, and each match votes for the  possible 3D poses of the object in an accumulator, in a Hough transform fashion.

\cite{2}  used this  method to  constrain a  SLAM system  by detecting  multiple
repetitive  object models.   They devised  a strategy  towards an  efficient GPU
implementation. Another  immediate industrial application is  bin picking, where
multiple instances  of the CAD model  is sought in a  pile of objects~\cite{21}.
The approach has also been used in robotic applications~\cite{20,22}.

In addition, several extensions to~\cite{1} have  been proposed.  A majority of  these works focused
on  augmenting   the  feature  description  to   incorporate  color~\cite{4}  or
visibility context~\cite{5}.   \cite{3} proposed  using points or  boundaries to
exploit  the  same  framework  in  order to  match  planar  industrial  objects.
\cite{23} modified the  pair description to include  image gradient information.
There are also  attempts to boost the accuracy and  performance of the matching,
without  modifying the  features. \cite{24}  made use  of the  special symmetric
object properties  to speed up  the detection  by reducing the  hash-table size.
\cite{25}  proposed a  scene specific  weighted  voting method  by learning  the
distinctiveness of the  features as well as the model  points using a structured
SVM.

Drost-PPF has often been criticized for  the high dimensionality of  the search
space~\cite{10},  for  its  inefficiency~\cite{7},  for being  sensitive  to  3D
background clutter  and sensor  noise~\cite{26}.  Furthermore,  other approaches
significantly  outperform it  on many  datasets~\cite{9,7}.  

%However,  these methods
%work with  RGB-D data and often are  worse at handling  occlusions.

In the  next section, we discuss in greater detail the shortcomings of~\cite{1} and propose
several suitable  feature sampling  strategies that allow  it to  outperform all
state-of-the-art methods~\cite{7,8,9,10} on the standard dataset of~\cite{7} and
the very  challenging occlusion dataset  of \cite{9},  in terms of recognition rate.
In addition, we show how we can speed up the approach to be significant faster than~\cite{1}.

%%%%%%%%%%%%%%%%%%%%%%%%%%%%%%%%%%%%%%%%%%%%%%%%%%%%%%%%%%%%%%%%%%%%%%%%%%%%%%%%

\section{Method}

We describe  here our contributions to  make PPFs more discriminative,  and more
robust  to background  clutter and  sensor noise.   We evaluate  the improvement
provided by each of these  contributions, and compare their combinations against
state-of-the-art methods in the next section.

\subsection{Pre-processing of the 3D Models and the Input Scene}

During a pre-processing stage, Drost-PPF subsamples  the 3D points of the target
objects and  the input  scene.  The  advantage is two-fold:  This speeds  up the
further  computations and  avoids considering too  many ambiguous  point pairs:
Points that  are close  to each other  tend to have  similar normals,  and
generate many non-discriminative PPFs. Drost-PPF therefore subsamples the points
so that two 3D points have at least a chosen minimal distance to each other.

This  however can  lead to a loss of useful  information when  normals are  actually
different. We therefore keep pairs even with a distance smaller than the minimal
distance if the  angle between the normals  is larger than 30  degrees, as these
pairs are likely to be discriminative. Subsampling is then done as in
Drost-PPF, but with this additional constraint.

\newcommand{\obj}{\text{obj}}

\subsection{Smart Sampling of Point Pairs}
\label{Sec:SmartSampling}
After sub-sampling, in  Drost-PPF, every scene point is paired  with every other
scene point during runtime.  The complexity is therefore quadratic in the number
of points in the 3D scene.   In order to reduce computation time, \cite{1}
suggests using only  every $m$-th scene point as the first  point, where $m$ is
often set  to 5 in  practice.  While  this improves runtime, the complexity remains
quadratic and  matching performance suffers  because we remove  information from
the already sampled scene point cloud.

We propose  a better way to  speed up the computations  without discarding scene
points: Given a first  point from the scene it should be  only paired with other
scene points that can  belong to the same object.  For  example, if the distance
between the two points is larger than the size of the object, we know that these
two points cannot possibly belong to the same object and therefore should not be
paired.  We  show below  that this  leads to  a method  that can  be implemented
much more efficiently.

A conservative  way to do so  would be  to ignore  any point that  is farther
away  than $d_\obj$  from the  first  point of  a  pair, where  $d_\obj$ be  the
diameter of the enclosing sphere of the target object, which defines a voting ball.

\begin{figure*}[t]
\begin{center}
\includegraphics[width=0.9\linewidth]{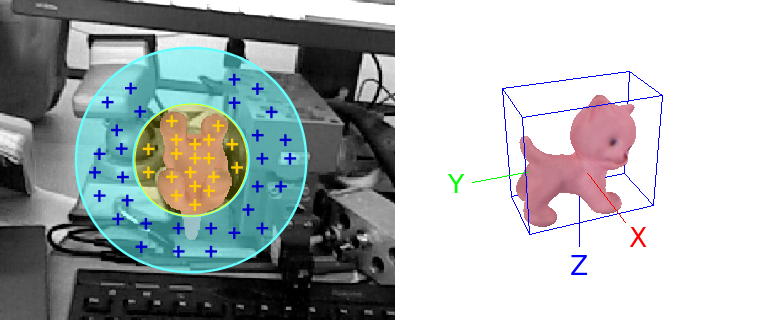}  \\
\end{center}
\caption{\label{fig:votingsphere}  Different voting  spheres for  a given  first
  point projected  into 2D.  The  yellow circle  corresponds to a  voting sphere
  with the smallest dimension of the object's bounding box as diameter. The blue
  circle  corresponds  to the  voting  sphere  with  the  same diameter  as  the
  enclosing sphere. Points sampled in the  yellow sphere are much more likely to
  lie on the object than points sampled in the blue sphere.}
\end{figure*}

However, a  spherical region can be  a very bad approximation  for some objects.
In particular, with narrow elongated objects, sampling from a sphere with radius
$d_\obj$ will  generate many points on  the background clutter if  the object is
observed in a  viewing direction parallel to its longest  dimension, as depicted
in Fig.~\ref{fig:votingsphere}.

In these cases we would like to use a smaller sampling volume where the ratio of
scene points lying on  the object compared to all other  scene points is larger.
However, we do not have any prior information  on the pose of the object and the
first scene point of  the pair can lie anywhere on the  object.  It is therefore
impossible to define a single volume that is smaller than the ball of radius $d_\obj$ without
discarding pairs  of scene  points under certain object configurations that  both lie  on the  target
object.

We  therefore opted  for using  consecutively  two voting  balls with  different
radiuses:   A   small  one   with   radius   $R_{\min}  =   \sqrt{d_{\min}^2   +
  d_\text{med}^2}$, where $d_{\min}$  is the smallest dimension  of the object's
bounding box and  $d_\text{med}$ is the median of its  three dimensions, and the
large conservative one  with radius $R_{\max} = d_\obj$.  It  can be easily seen
that $R_{min}$ is the smallest observable  expansion of the object.  We will say
that a  point pair is  accepted by a  voting ball if the  first point is  at the
center of  the ball and  its distance  to the second  point is smaller  than the
radius of the ball.

%% the length of the  diagonal of the smallest  rectangle of the
%% bounding  box  and is  the  diameter  of the  sphere  that  covers the  smallest
%% observable expansion of the object.

We first populate the accumulator with votes from pairs that are accepted by the
small ball.  We  extract the peaks from the accumulator,  which each corresponds
to a hypothesis on  the object's 3D pose and correspondence  between a model and
scene point, as in Drost-PPF.  We  then continue populating the accumulator with
votes from  pairs accepted  by the  large ball  but were  rejected by  the small
one. We  proceed as  before and  extract the  peaks to  generate pose  and point
correspondence hypotheses.  This way, under poses such as the one illustrated by
Fig.~\ref{fig:votingsphere},  we  can  get  peaks  that  are  less  polluted  by
background clutter  during the  first pass,  and still get  peaks for  the other
configurations during the second pass.

To efficiently look for pairs accepted by a ball of radius $d$, we use a spatial
lookup table:  For a given scene,  we build a  voxel grid filled with  the scene
points.  The number of  voxels in each dimension is adapted  to the scene points
and can  differ in $x$, $y$,  and $z$ dimensions.   Each voxel in this  grid has
size  $d$,  and  stores  the  indices  of the  scene  points  that  lie  in  it.
Reciprocally, for  each scene point  we also store the  voxel index to  which it
belongs.   Building up  this voxel  grid  is a  $O(n)$ operation.   In order  to
extract for a  given first point all  other scene points that  are maximally $d$
away,  we first  look up  the voxel  the scene  reference point  belongs to  and
extract all  the scene point  indices stored in this  voxel and in  all adjacent
voxels.   We check  each of  these scene  points if  its distance  to the  scene
reference point is actually smaller or equal than $d$.

The complexity of this method is therefore $O(nk)$ where $k$ is usually at least
one  magnitude  smaller  than  $n$,  compared to  the  quadratic  complexity  of
Drost-PPF, while guaranteeing that all relevant point pairs are considered.

\subsection{Accounting for Sensor Noise when Voting}
\label{sec:noise}
For fast access, the PPFs are  discretized. However, sensor noise can change the
discretization bin, preventing  some PPFs from being correctly  matched.  We overcome
this  problem  by  spreading  the  content  of  the  look-up  table  during  the
pre-processing of  the model. Instead of  storing the first model  point and the
rotation angles  only in the  bin indexed by the  discretized PPF vector,  we also
store them  in the (80)  neighborhood bins indexed by the adjacent discretized PPFs 
(there are $3^4$  = 81 -  1 adjacent bins).

We face a  similar problem at run-time during voting  for the quantized rotation
angles  around the point normals.  To  overcome this,  we  use the  same
strategy as  above and vote not  only for the original  quantized rotation angle
but also for its adjacent neighbors.

%% While spreading is a powerful tool, we have to make sure
%% that the quantized bins are not too large. Too large bins in the context of spreading
%% easily lead to too much generalization which might reduce the discriminative power of 
%% PPFs and lead to less robust detection results. 
%% Therefore, we reduce the bins compared to~\cite{1}. In practice, 
%% we quantize  each angle of Eq.~\eqref{eq:PPF} into $8.2^{\circ}$ steps and the distance 
%% into steps of $\frac{1}{40}d_\obj$. The rotation space in the accumulator space is
%% quantized into $11.25^\circ$ steps. 

%% In order  not to  choose too  large discretization steps  and to  generalize too
%% much, we  use smaller discretization  levels than  Drost-PPF. We cover  at least
%% $\pm 8$ degrees  of normal angle noise, and our  distance discretization step is
%% 0.025 of  the object  diameter, thus,  we are robust  to at  least $\pm  0.025 *
%% d_\obj$ in  distance measure  noise.  We  also do  the same  when voting  for an
%% in-plane rotation  where we do not  only vote for the  corresponding discretizdd
%% in-plane angle but also for its adjacent neighbors.

\begin{figure*}[t]
\begin{center}
\includegraphics[width=0.9\linewidth]{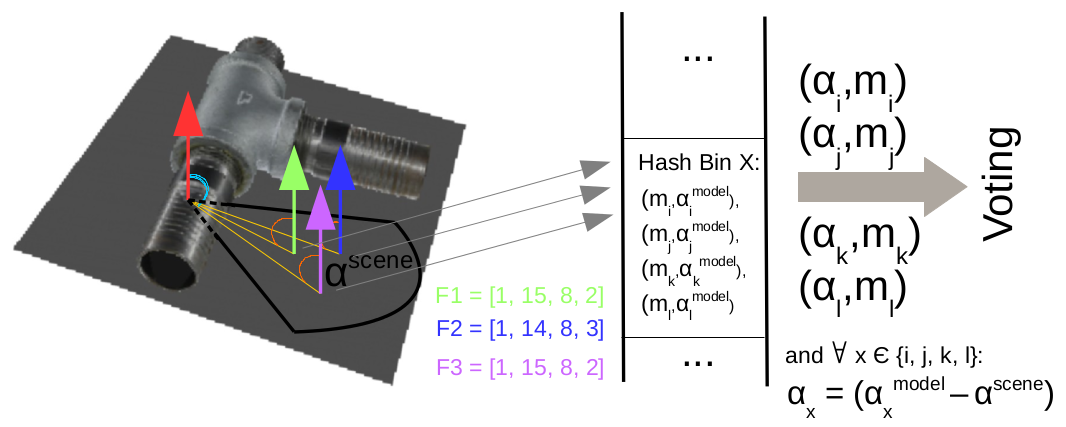}  \\
\end{center}
\caption{\label{fig:generalization} A first scene point, in red, and three other
  scene  points  (green, blue  and  purple)  build  three point  pair  features.
  Because  of  discretization  and  spreading, all  three  point  pair  features
  correspond  to  the  same scene  rotation  angle  $\alpha^{scene}$.   Furthermore,  feature
  discretization of {\sf F1} and {\sf F3}  and spreading of {\sf F2} might lead to
  all three features being mapped to the  same hash bin {\sf X} which results in
  voting for the  same combinations of model reference points and rotation angles 
  $({\sf m}_{\sf i}, \alpha_i)$, $({\sf m}_{\sf j}, \alpha_j)$, 
  $({\sf m}_{\sf k}, \alpha_k)$, $({\sf m}_{\sf l}, \alpha_l)$ several times.
  This artificially increases the voting  for these specific combinations of model
  points and rotation  angles and deteriorates the performance, especially in cases as shown in this figure where votes come from background.}
\end{figure*}

However,  as  shown  in  Fig.~\ref{fig:generalization},  spreading  also  has  a
drawback: Because of discretization and spreading  in the feature+rotation space, it
is very likely  that pairs made of  close scene points have the same quantized rotation angle 
and are mapped  to the same look-up table bin. They will thus vote  for the same bin
in the accumulator space, introducing a bias in the votes.

A direct  method to  avoid multiple  votes would be  to use  a 3D  binary array
$(a_{i,j,k})$ for  each scene point  to ``flag'' if  a vote  with the
$i$-th model point,  $j$-th model point as first and  second point respectively,
and $k$-th quantized rotation angle around the normal has already been cast, and
prevent additional votes for this combination.

Unfortunately,  such an array   would  be very  large as  its size  is
quadratic with the number  of model points, and we would need  to create one for
each of the scene point.  We propose a much more tractable solution.

%% A na\"ive implementation would  require a
%% very large amount of  memory and the number of CPU  cache misses would increase,
%% slowing down  the computations.  We  tried storing  these arrays as  matrices of
%% 32-bit integers, where each bit corresponds to one discretized in-plane rotation
%% angle, but in  practice the method becomes inefficient when  the number of model
%% points is larger than 650.

%% Therefore, we propose an alternative solution where the size of the consistency matrix
%% does not scale quadratically to the number of model points.

%% Instead, it is constant in size for all objects and scales linearly 
%% with the number of all possible quantizations of the PPFs. In practice, we
%% quantize each angle and the distance of Eq.~\eqref{eq:PPF} into 22 and 40 bins
%% respectively, yielding to $22^3 \times 40$ possible quantized PPFs.

%% $\bF(\bm_1, \bm_2, \bn_1, \bn_2)$ (in practice, 
%% we quantize each of $\angle(\bn_1, \bd), \angle(\bn_2, \bd), \angle(\bn_1, \bn_2) \in \left[0^{\circ}...180^{\circ}\right]$ 
%% into $22$ bins and $\|\bd\|_2 \in \left[0...d_\obj\right]$ into $40$ bins resulting in
%% $22^3 \times 40$ possible quantized point pair features).

  Instead  of  indexing  a  flag  array  by  the  pair  of model points  and
  corresponding rotation, we use an array  $b$ indexed by the quantized PPFs the
  point pairs generate.  Each element of  $b$ is a 32-bit integer initialized to
  0, and  each bit corresponds to  a discretized rotation angle $\alpha^{scene}$.   We simply set
  the bit corresponding to  a quantized PPF and scene rotation to 1  the first time it
  votes, and to prevent further voting for the same combination of quantized PPF
  and scene rotation, even  if it is generated again by  discretization and spreading.
  Spreading  is achieved  by  generating  for each  discretized  PPF created  at
  run-time from  the scene  points the adjacent  discretized PPFs,  and treating
  them as the original discretized PPFs.

%% To introduce spreading, we vary for each PPF  its quantized values to generate
%% adjacent PPFs that also vote.

%% Note that we have to use another rotation than the one we used in $(a_{i,j,k})$.
%% This is  because the  rotation used  in $(a_{i,j,k})$ is  dependent on  both the
%% scene and the model  point pair and thus, is usually  different for each element
%% in  the look-up  table  paired  with the  current  scene  point pair.   However,
%% \cite{1} showed that such  rotation can be split into a scene  and a model part,
%% where the model part is constant at runtime and the scene part is only dependent
%% on the scene  point pair.  Thus, it is  the same for all elements  stored in one
%% bin in the look-up table and allows us to leverage it to perform flagging in $b$
%% for look-up table bins.

  %% \vincentrmk{the rotation thing is still not clear, we should do something
  %%   about it:}
  
Note that we use here the scene rotations around the normals since it is only dependent on the
scene point pair and not on the model points as shown in~\cite{1}. Thus, it is the same for
all elements stored in one bin in the look-up table which allows us to leverage it
to perform flagging in $b$.

%are dependent on both  the model and
%the scene as shown in Sec.~\ref{DrostSec} and Fig.~\ref{fig:generalization}.
%However, \cite{1}  showed that  such rotation can  be split  into a
%scene and  a model part,  where the  model part is  constant at runtime  and the
%scene part is only dependent on the scene  point pair.  Thus, it is the same for
%all elements stored in one bin in the look-up table and allows us to leverage it
%to perform flagging in $b$.

%% In this context we have to mention that  we can not use the same rotation angles
%% for flagging that we  would have used in the na\"ive  approach.  This is because
%% they  are dependent  on both  the model  and the  scene, and  thus, are  usually
%% different  for  the single  elements  stored  in a  bin  in  the look-up  table.
%% However,~\cite{1} has  shown that each such  rotation can be split  into a
%% scene and  a model part,  where the  model part is  constant at runtime  and the
%% scene part is only dependent on the scene  point pair.  Thus, it is the same for
%% all elements stored in one bin in the look-up table and allows us to leverage it
%% to perform flagging in $b$.

This solution is  more efficient than the direct method  discussed above, as the
number of possible entries is smaller in practice, thanks to the quantization of
the PPFs.  $b$ is constant in size  for all objects and scales linearly with the
number of all possible quantizations of the PPFs.  In practice, we quantize each
angle and the  distance of Eq.~\eqref{eq:PPF} into 22 and  40 bins respectively,
yielding to  $22^3 \times 40$  possible quantized PPFs.  In  our implementation,
when  the  number  of  model  points   exceeds  650,  the  direct  method  takes
significantly more  time, with a  typical slow-down of  factor 3 for  1000 model
points.
%% Typical speed ups originated from this technique are provided in the
%%   experiment section. 
%\stefanrmk{I can't generate exact numbers any
%    more since the old code is not up to date. what should we do about
%    it?}\vincentrmk{can you just give a rough estimate from what you remember?}
%% To give an order of magnitude, in case of an object with 1000 model points, 
%% \vincentrmk{how much is it?
%%   Compare to the size of $a$}
%% \stefanrmk{this depends on the number of model points. In case we have an object with 
%% 1000 model points the difference is quite drastic - in case of 650 model points there is no difference. 
%% But also note, that we check at look-up table level now: that means we flag a WHOLE bin and thus, allow allow elements to vote or none of them.
%% So flagging is done at a much higher level and not at element level as before. 
%% This also brings quite some speed-up. Should we mention that?}

\subsection{Generating Object Pose and Postprocessing}

To extract object poses from the accumulator, Drost-PPF uses a greedy clustering
approach.  They extract the peaks from  the accumulator, each corresponding to a
hypothesis on the object’s 3D pose  and correspondence between a model and scene
point, and process them in the same  order as their numbers of votes, and assign
them to the closest  cluster if it is close enough,  or otherwise create another
cluster.

We found  that this method  is not always reliable  especially in case  of noisy
sensors and background clutter. These result in spurious votes and the number of
votes in the  accumulator space is not necessarily a  reliable indicator for the
quality of a hypothesis.

% Vincent: I removed this to avoid making your voting method uselessly guilty:

%% In case of two voting balls where  peaks in the accumulator space of the smaller
%% voting ball tend to  have less votes than the ones from  the larger voting ball,
%% this observation becomes very clear.

%% in  which hypotheses  originating  from different  voting  balls are  treated
%% separately.

%%  and runs a where each hypothesis makes  up a cluster with its pose being the
%%  cluster pose

% That's bottom-up clustering:

%%  Each cluster then searches through  all neighboring hypotheses and adds them
%%  to the cluster if their poses are similar enough to the cluster pose.

Therefore, we propose  a different cluster strategy that takes  into account our
voting  strategy.  We  perform a  bottom-up  clustering of  the pose  hypotheses
generated  during voting  with one  voting ball.   We allow  hypotheses to  join
several clusters as  long as their poses  are similar to the one  of the cluster
center. We also  keep track of the model points  associated with each hypothesis
and only allow  a hypothesis to vote  for a cluster if no  other hypothesis with
the same  model point has  voted for this cluster  before.  Thus, we  avoid that
ambiguous and repetitive geometric structures  such as planar surfaces introduce
biases.

For each  of the  few first  clusters with  the largest  weights, we  refine the
estimated  pose using  projective  ICP~\cite{projective_icp}.   In practice,  we
consider the four first clusters for each of the two voting balls.

To reject the clusters  that do not actually correspond to  an object, we render
the object  according to the  corresponding 3D pose,  and count how  many pixels
have a depth close  to the one of the rendering, how many  are further away from
the camera---and could be occluded, and  how many are closer---and are therefore
not consistent with the rendering. If the number of pixels that are closer is
too large compared to the total number of pixels, we reject the cluster.

In practice, this  threshold is set to  10\%. We also discard  objects which are
too much occluded. As  a last check, we compute areas  with significant depth or
normal change in the scene, and compare  them to the silhouette of the projected
object: If the  silhouette is not covered  enough by depth or  normal change, we
discard the  match.  In  practice, we  use the  same threshold  that we  use for
occlusion check.  We  finally rank all remaining clusters according  to how well
they fit the scene  points and return the pose of the best  one only, or in case
of  multi-instance  detection,  the  whole  list of  poses  from  all  remaining
clusters.

\section{Experimental Results}

We compare  here our method against  Drost-PPF, Linemod~\cite{7}, DTTs~\cite{8},
Birdal   et    al.~\cite{6},   Bachmann   et   al.~\cite{10},    and   Krull   et
al.~\cite{9} on the ACCV dataset of~\cite{7} and on the Occlusion Dataset
of~\cite{9}.

For our method, we use the same parameters for all the experiments and objects,
except for the post processing thresholds to account for the specificities of
the occlusion dataset.

%% For  all  our  experiments  and  objects, we  have  not  changed  our  detection
%% parameters  and have  used the  same set  of parameters  across experiments  and
%% objects.  However, for  the occlusion  dataset, we  adapted the  
%% thresholds to account for the large number of occlusion (60\%).

\subsection{ACCV dataset of~\cite{7}}

We  first tested  our  method on  the standard  benchmark  from \cite{7}.   This
dataset contains  16 objects  with over  1100 depth and  color images  each, and
provides the ground truth 3D poses for each object.  We only evaluate our method
for the non-ambiguous objects ---we removed  the bowl and the cup--- because
state-of-the-art approaches considered only these objects.

\cite{6} does  not evaluate all  objects and does  not use refinement  with ICP.
For the approach  of Bachmann et al.~\cite{10}, we use  the numbers reported for
synthetic training  without a ground plane,  since adding a ground  plane during
training artificially adds additional knowledge about the test set.
\begin{table*}
\begin{center}
\ra{1.3}
\small
\begin{tabular}{@{}lcccccc@{}}
\toprule
Approach & Our Appr. & Linemod~\cite{7} & Drost~\cite{1} & DTT~\cite{8} & Brachmann~\cite{10} & Birdal~\cite{Birdal2015} \\
\midrule
Sequence (\#pics) & \multicolumn{6}{c}{Matching Score} \\
\midrule
Ape (1235)        & {\bf98.5\%} &      95.8\% &	 86.5\% &      95.0\% &	     85.4\% & 81.95\%\\
Bench V. (1214)   & {\bf99.8\%} &      98.7\% &	 70.7\% &      98.9\% &	     98.9\% &  --    \\
Cam (1200)        & {\bf99.3\%} &      97.5\% &	 78.6\% &      98.2\% &	     92.1\% & 91.00\%\\
Can (1195)        & {\bf98.7\%} &      95.4\% &	 80.2\% &      96.3\% &	     84.4\% &  --    \\
Cat (1178)        & {\bf99.9\%} &      99.3\% &	 85.4\% &      99.1\% &      90.6\% & 95.76\%\\
Driller (1187)    &     93.4\%  &      93.6\% &	 87.3\% &      94.3\% & {\bf99.7\%} & 81.22\%\\
Duck (1253)	  & {\bf98.2\%} &      95.9\% &	 46.0\% &      94.2\% &	     92.7\% &  --    \\
Box (1253)        &      98.8\% & {\bf99.8\%} &  97.0\% & {\bf99.8\%} &      91.1\% &  --    \\
Glue (1219)	  &      75.4\% &      91.8\% &	 57.2\% & {\bf96.3\%} &	     87.9\% &  --    \\
Hole P. (1236)    & {\bf98.1\%} &      95.9\% &	 77.4\% &      97.5\% &	     97.9\% &  --    \\
Iron (1151)       &      98.3\% &      97.5\% &	 84.9\% &      98.4\% &	{\bf98.8\%} & 93.92\%\\
Lamp (1226)       &      96.0\% &      97.7\% &	 93.3\% & {\bf97.9\%} &	     97.6\% &  --    \\
Phone (1224)      & {\bf98.6\%} &      93.3\% &	 80.7\% &      88.3\% &	     86.1\% &  --    \\
\bottomrule
\end{tabular}
\end{center}
\caption{\label{tab:exp1} Recognition  rates according to the  evaluation metric
  of~\cite{7} for different methods.  We perform  best for eight out of thirteen
  objects, while \cite{7,10,8} use color data  in addition to depth, and we only
  use depth data.}
\end{table*}

Like~\cite{10,8,7} we  only use  features that
are visible  from the upper hemisphere  of the object.  However,  differently to
these methods and similarly  to~\cite{1,Birdal2015} we still allow the
object to be  detected in any arbitrary  pose.  Thus, we are solving  for a much
larger search space than~\cite{10,8,7}.

In addition, as in~\cite{1,Birdal2015}, we only use the depth information and do
not make use of the color  information as~\cite{10,8,7}.  However, as shown
in Table~\ref{tab:exp1}, we perform best on eight objects out of thirteen.

\subsection{Occlusion Dataset of~\cite{9}}

The  dataset of~\cite{7}  is almost  free of  occlusion.  Hence,  to demonstrate
robustness with  respect to occlusions,  we tested  our method on  the occlusion
dataset of \cite{9}. It is much  more noisy and includes more background clutter
than previous  occlusion datasets~\cite{16}.  It  includes over 1200  real depth
and color images with 8 objects and their ground truth 3D poses.

As Table~\ref{tab:exp2}  shows, our  approach performs  better for  five objects
while the method of Krull et  al.~\cite{9} performs better on three objects.  On
average we perform 3.3\% better than Krull that was the state-of-the-art on this
dataset.  While we  only  use depth  data, \cite{9}  also  uses color  data.
Moreover, the method in~\cite{9} was trained with an added a ground plane during
training\footnote{Private email  exchange with the authors.},  which gives an extra
advantage.
\begin{table*}
\begin{center}
\ra{1.3}
\small
\begin{tabular}{@{}lcccc@{}}
\toprule
Approach & Our Approach & Linemod~\cite{7} & Brachmann~\cite{10} & Krull~\cite{9} \\
\midrule
Sequence & \multicolumn{4}{c}{Matching Score} \\
\midrule
Ape          & {\bf81.4\%} &      49.8\% &	 62.6\% &      77.9\% \\
Can          & {\bf94.7\%} &      51.2\% &	 80.2\% &      86.6\% \\
Cat          &     55.2\%  &      34.9\% &	 50.0\% & {\bf55.6\%} \\
Driller      &     86.0\%  &      59.6\% &	 84.3\% & {\bf93.6\%} \\
Duck 	     & {\bf79.7\%} &      65.1\% &	 67.6\% &      71.9\% \\
Box          & {\bf65.6\%} &      39.6\% &        8.5\% &      35.6\% \\
Glue  	     &      52.1\% &      23.3\% &	 62.8\% & {\bf67.9\%} \\
Hole Puncher & {\bf95.5\%} &      67.2\% &	 89.9\% &      94.8\% \\
\midrule
Average	     & {\bf76.3\%} &      48.8\% &       63.2\% &      73.6\% \\
\bottomrule
\end{tabular}
\end{center}
\caption{\label{tab:exp2}     Recognition     rates     on     the     occlusion
  dataset~\cite{9}     according     to    the     evaluation     metric
  of~\cite{9}.  Our  approach performs  better for  five objects  out of
  eight.   On  average  we  perform 3.3\%  better  than  Krull~\cite{9}, while
  Krull~\cite{9} uses color data in addition to depth and a ground plane
  during training. We only  use depth data. }
\end{table*}
%% Fig 2) Results on the occlusion dataset of \cite{9}. Our approach performs better for
%% five objects  while the  method of  Krull et  al. \cite{9}  performs better  on three
%% objects. In  average we perform 3.3\%  better than Krull, although,  contrary to
%% Krull et al \cite{9}, we only make use of depth data.

\subsection{Computation Times}

Our approach  takes between $0.1s$  and $0.8s$  to process a  $640\times480$ depth
map.  Like Drost-PPF,  the  sub-sampling during  pre-processing  depends on  the
object diameter;  denser sampling is  used for smaller objects,  which increases
the processing time.

On average  it is  about  6 times  faster than  Drost-PPF, while  being
significantly more accurate.  Moreover, our method could be  implemented on GPU,
as \cite{2} did, for further acceleration.

%% \stefan{Note that different runtimes originate from the different objects used,
%% since similar to \cite{1}, our point sampling is dependent on the object diameter $d_\obj$.
%% Thus, smaller objects usually lead to a denser sampling of the scene point cloud and therefore to 
%% a longer processing times.}
%% As we see in Fig.~\ref{tab:exp3}, our approach is significantly more efficient
%% than the method of  Drost et al~\cite{1}. This is even  more true, since we
%% take every  scene reference  point into account  while~\cite{1}  uses every
%% $n$-th scene point (in practice, $n = 5$).

\subsection{Worst Case Runtime Discussion}

In this section we discuss the worst case runtime behavior. Despite our smart sampling strategy presented in Sec.~\ref{Sec:SmartSampling}, the worst case runtime behavior is still $O(n^2)$ where $n$ is the number of subsampled points in a scene. 
However, this is not a problem in practice as this happens only if all observed scene points lie in a sphere with radius $\sqrt{3} \times d_{obj}$ ($\sqrt{3}$ comes from taking the diagonal of voxels of width $d_{obj}$ into account). This is because our spatial look-up table only gives back points that fall into the same voxel or adjacent voxels of the spatial look-up table as our candidate point falls into. When there are points outside this sphere, these points do not vote for this candidate point thus resulting in a runtime $O(nk)$ with $k < n$.

\begin{figure*}
\begin{center}
\begin{tabular}{ccc}
\includegraphics[width=0.5\linewidth]{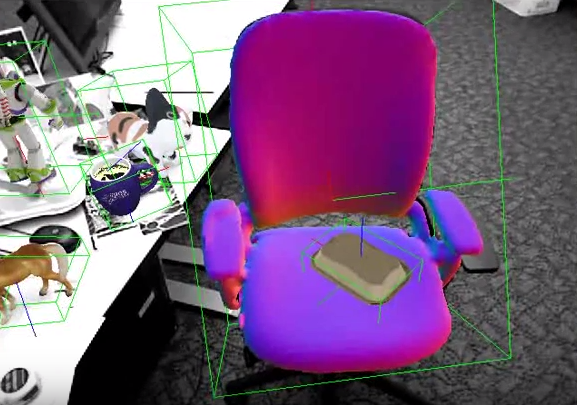} &
\includegraphics[width=0.5\linewidth]{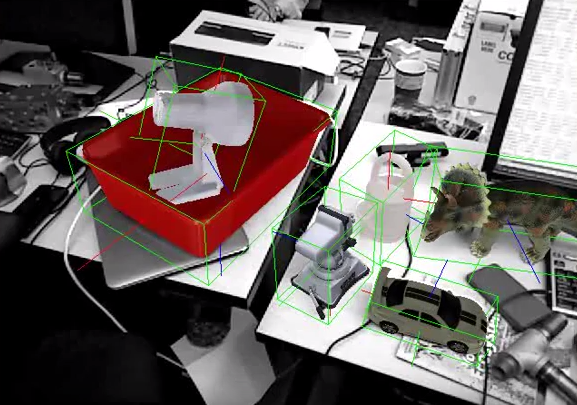} &
\end{tabular}
\end{center}
\caption{\label{fig:teaser2}  Several 3D objects are
simultaneously  detected with our method under  different  poses on  cluttered
background  with  partial occlusion  and  illumination changes.   Each
detected  object is augmented  with its  3D model, its 3D bounding box and its
coordinate systems. For better visibility, the background is kept in gray.}
\end{figure*}

Taking normals into account for subsampling increases the number of subsampled points by a factor typically smaller than two. Therefore the number of points falling into one voxel of the spatial look-up table is not very large and due to (self-) occlusion the overall number of visble points is fairly small, typically around 1000 or less. However, even 1000 points are easily handled in a $O(n^2)$ manner and in such a case matching is done quite quickly. For instance, matching for a close-up view of the chair seen in Fig.~\ref{fig:teaser2} is usually done in less than 200ms.

More problematic are scenes where subsampled points on an object are only a tiny fraction of the overall number of points (the overall number of points can easily exceed $15k$). However, in these cases our spatial look-up table kicks in and the run-time goes from $O(n^2)$ for Drost et al.~\cite{1} to $O(kn)$ where $k << n$. $k$ is often over twenty times smaller than $n$.

In short, in practice, the complexity of our approach is significantly better than Drost's~\cite{1}.

\subsection{Contribution of each Step}

We also performed experiments on the occlusion dataset~\cite{9} with all
eight objects  to evaluate the  influence on the matching  score of each  of the
steps  we proposed  compared  to~\cite{1}.

\newcommand{\Drost}{\text{Drost}}
\newcommand{\Ours}{\text{Ours}}

To do so, we first ran our implementation of the original Drost-PPF method, and
computed the average matching score $\overline{S_\Drost}$ for all eight objects on the dataset of~\cite{9}.  We
then turned on all our contributions and computed the new average matching score
$\overline{S_\Ours}$. The gain $g_c$ from each contribution alone is computed by
computing the average matching score $\overline{S_c}$ with only this contribution turned
on, and taking $g_c = \overline{S_c} / (\overline{S_\Ours} - \overline{S_\Drost})$.

\begin{figure*}
\begin{center}
\begin{tabular}{cc}
\includegraphics[width=0.5\linewidth]{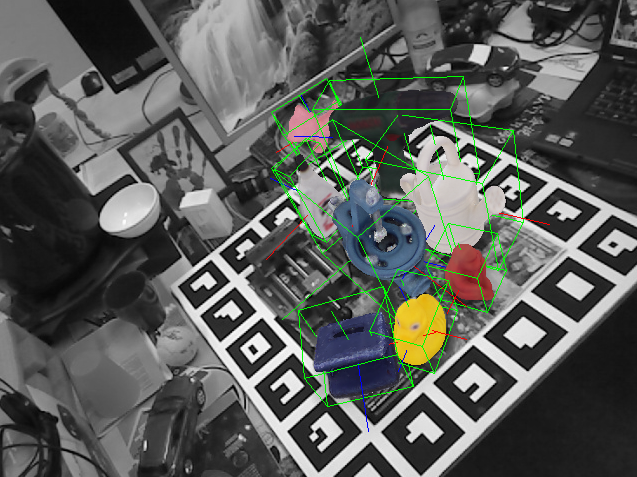} &
\includegraphics[width=0.5\linewidth]{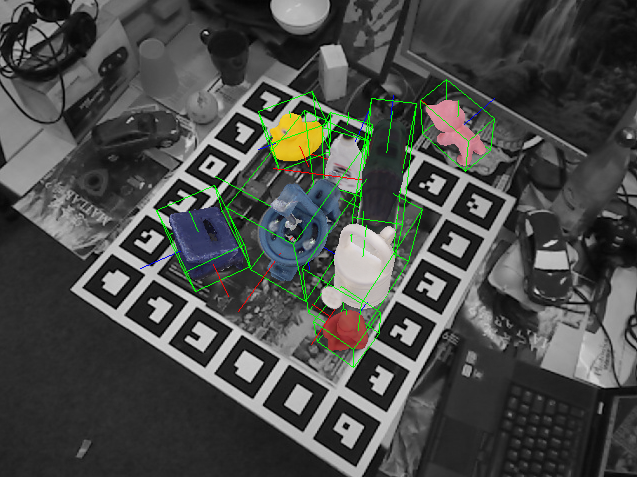} \\
\includegraphics[width=0.5\linewidth]{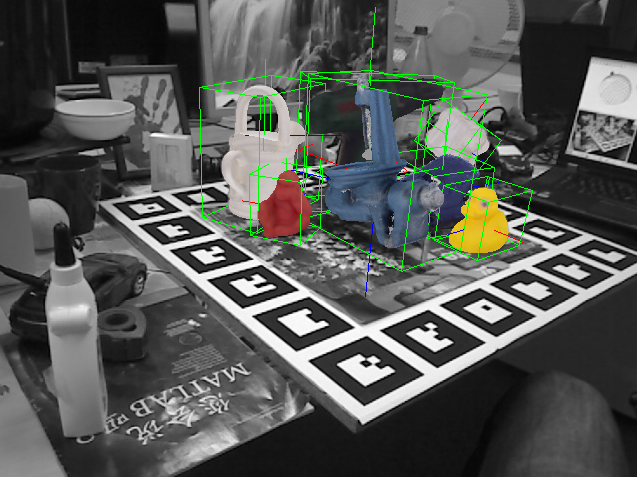} &
\includegraphics[width=0.5\linewidth]{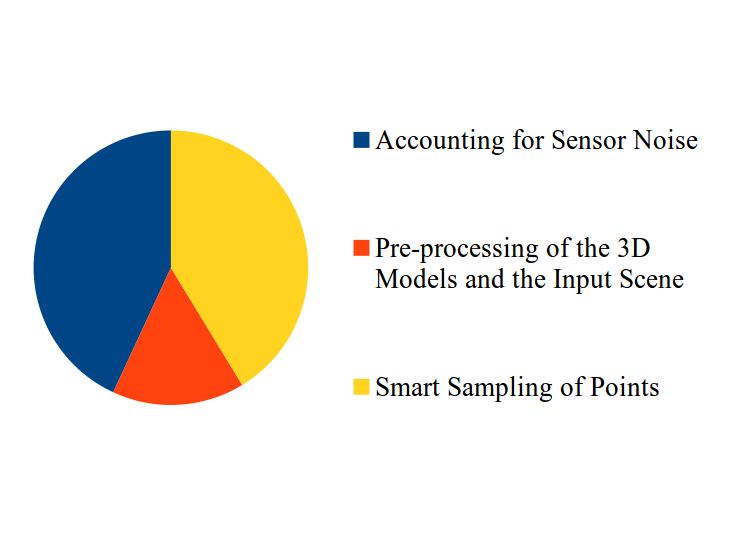} 
\end{tabular}
\end{center}
\caption{\label{fig:influence} Three images of the dataset of~\cite{9}
  with at least seven out of nine objects correctly detected. Note the strong
  background clutter and heavy occlusion. 
  Lower Right: Contribution of  each proposed
  step to the matching scores compared  to \cite{1}. Accounting for sensor noise
  and our sampling of points are almost equal, while our proposed pre-processing
  of the  3D model  and the  input  scene has  the smallest  but still  significant
  effect.}
\end{figure*}

As Fig.~\ref{fig:influence} shows, accounting for  sensor noise is with $43.1\%$
the most  important part,  directly followed  by smart  sampling of  points with
$41.3\%$ and finally our contribution to  pre-processing of the 3D model and the
input scene by $15.6\%$.
%% \vincentrmk{How is this computed exactly?}
%% \stefanrmk{I switched off the single contributions and computed the average matching score. This avg. 
%% matching score was substracted from the maximum avg. score achieved by our method where all contributions are
%% switched on. Then I added all the differences and divided each of the single differences by the sum of differences.
%% Not sure if we can say exactly what I said in the text given how I did the experiments.}

\section{Conclusion}

We have shown that by cleverly sampling features and by adding feature spreading
to account for  sensor noise, we can boost the  method from Drost~\cite{1}
to outperform state-of-the-art approaches in  object instance detection and pose
estimation, including those that use additional information such as color cues.

%% We
%% have included  extensive experiments  on state-of-the-art  datasets~\cite{7,9} that
%% show this superiority.

%%%%%%%%%%%%%%%%%%%%%%%%%%%%%%%%%%%%%%%%%%%%%%%%%%%%%%%%%%%%%%%%%%%%%%%%%%%%%%%%

\bibliographystyle{splncs03}
\bibliography{./string,./vision}

\end{document}